%% file: main.tex
\documentclass{article}
\usepackage{PRIMEarxiv}
\usepackage[utf8]{inputenc}
\usepackage[T1]{fontenc}
\usepackage{multirow}
\usepackage{lipsum}
\usepackage{graphicx}
\usepackage{tikz}
\usepackage{pgfplots}
\usepackage{amsfonts}
\usepackage{multirow} 
\usepackage{pgfplots}
\usepackage{amsmath}
\usepackage{threeparttable}
\usepackage{amsmath}
\usepackage{amssymb}
\usepackage{booktabs}
\usepackage{stfloats}
\usepackage{cuted}
\pgfplotsset{compat=1.18} 
\usepackage{algorithm}
\usepackage{algpseudocode}
\usepackage{amsmath}
\usepackage[numbers]{natbib}      
\newcommand{\Input}{\item[\textbf{Input:}]}
\newcommand{\Output}{\item[\textbf{Output:}]}

\title{Adapting to the Unknown: Robust Meta-Learning for Zero-Shot Financial Time Series Forecasting}

\author{
  Anxian Liu$^{1}$\thanks{Equal contribution.} \quad
  Junying Ma$^{1}$\footnotemark[1] \quad
  Guang Zhang$^{2}$\thanks{Corresponding author.} \\
  $^{1,2}$The Hong Kong University of Science and Technology (Guangzhou) \\
  $^{1}$\{aliu789, jma652\}@connect.hkust-gz.edu.cn \\
  $^{2}$guangzhang@hkust-gz.edu.cn
}

\begin{document}
\maketitle

\begin{abstract}
Financial time series forecasting in zero-shot settings is critical for investment decisions, especially during abrupt market regime shifts or in emerging markets with limited historical data. While Model-Agnostic Meta-Learning (MAML) approaches show promise, existing meta-task construction strategies often yield suboptimal performance for highly turbulent financial series. To address this, we propose a novel task-construction method that leverages learned embeddings for both meta task and also downstream predictions, enabling effective zero-shot meta-learning. Specifically, we use Gaussian Mixture Models (GMMs) to softly cluster embeddings, constructing two complementary meta-task types: intra-cluster tasks and inter-cluster tasks. By assigning embeddings to multiple latent regimes probabilistically, GMMs enable richer, more diverse meta-learning. This dual approach ensures the model can quickly adapt to local patterns while simultaneously capturing invariant cross-series features.  Furthermore, we enhance inter-cluster generalization through hard task mining, which identifies robust patterns across divergent market regimes. Our method was validated using real-world financial data from high-volatility periods and multiple international markets (including emerging markets). The results demonstrate significant out-performance over existing approaches and stronger generalization in zero-shot scenarios.
\end{abstract}

\input{1_intro}
\input{2_related}
\input{3_method}

\input{4_experiment}

\input{5_conclusion}

\newpage
\bibliographystyle{plainnat}
\bibliography{main}  

\end{document}

%% file: 1_intro.tex
\section{Introduction}
Black Swan events—extreme financial outliers with catastrophic consequences, haunts modern stock markets. Paradigm-shifting moments like the 1987 Black Monday crash (22.6\% single-day decline) and Long-Term Capital Management's 1998 collapse reveal the fundamental fragility of traditional market assumptions \cite{cont2001empirical}. Traditional models trained on historical data frequently underperform during significant market shifts \cite{aldoseri2023re}. The COVID-19 crisis in March 2020 crystallized this vulnerability: the Dow Jones Industrial Average (DJIA) plunged 6,400 points (about 26\%) in just four trading days. Traditional forecasting models failed to capture the nonlinear effects of public health emergencies \cite{moghar2020stock}, particularly given that epidemics represent extreme tail events with devastating risk potential \cite{zhang2022transformer}.

This financial cognitive dissonance stems from a market microstructure paradox: while price trajectories are path-dependent, the underlying mechanisms of volatility clustering and liquidity shocks exhibit fractal recurrence across regimes \cite{cont2001empirical}. Deep learning models, particularly Long Short-Term Memory (LSTM), Gated Recurrent Unit (GRU), and Transformers, have demonstrated superior performance in financial time series analysis \cite{moghar2020stock,shen2018deep,zhang2022transformer}. However, their successful training typically requires large-scale datasets, limiting applicability during sudden market changes when data availability is constrained and relevant historical crisis data is scarce. This challenge is especially pronounced in emerging markets, which exhibit weaker post-crisis recovery resilience compared to developed markets \cite{patel1998crises}.
 
Recent advances in few-shot learning have shown promise in addressing these challenges. \citet{Wood_2024} demonstrated that few-shot learning techniques can achieve an 18.9\% improvement in risk-adjusted returns during regime shifts by capturing market patterns with limited historical data. Existing solutions can be categorized into three approaches: data-based \cite{zhang2023few}, metric-based \cite{wang2021metric}, and optimization-based \cite{narwariya2020meta}. However, financial time series present unique challenges for few-shot learning due to high volatility and non-stationarity. These challenges include: (1) The requirement for carefully curated support sets during rapid market shifts; (2) The sensitivity to support example selection, which introduces potential bias; and (3) The inability to handle unprecedented events with no historical patterns \cite{li2024survey}.

Emerging zero-shot learning approaches leveraging fundamental market principles and cross-asset relationships offer promising alternatives for novel situations \cite{Wood_2024}. While specific market events may be unprecedented, the underlying economic relationships and market mechanics often follow discoverable patterns. These approaches exploit the inherent relationships between different financial instruments and market indicators to generate predictions for unprecedented market conditions without requiring direct historical examples.

To address the dual challenges of limited historical data and non-stationary market dynamics, we propose a novel \textbf{MAML-based meta-learning framework} tailored to financial time series in the zero-shot scenario. Our contributions are as follows:

\begin{itemize}
    \item We propose a dual-purpose encoder that generates embeddings for both meta task construction and downstream predictions. During meta training, the encoder first produces embeddings to guide task construction, and these task-aware representations are then utilized for downstream predictions.
    \item We design a two-level task construction strategy based on the encoder outputs: (1) Intra-cluster tasks that group sequences within the same cluster to encourage local consistency and the effectiveness of adaptation; (2) Inter-cluster tasks comprise regular tasks with randomly paired components and hard negative tasks constructed from component pairs with maximum statistical distance. This design enforces the learning of distribution-invariant features essential for robust cross-domain adaptation.
    \item We evaluate our framework on stock market datasets from high-volatility periods and multiple international markets, including constituent stocks of S\&P 500, N225, HSCI, and NSE. The model is trained on a subset of stocks from each market and tested on the remaining unseen constituents within the same market. We demonstrate that our proposed framework achieves strong performance in this challenging zero-shot cross-stock prediction scenario.

\end{itemize}

%% file: 2_related.tex
\section{Related Work}
Zero-shot learning (ZSL) aims to enable machine learning models to recognize classes not seen during training by leveraging auxiliary semantic information. 

Meta learning, or "learning to learn," is one of the important categories of ZSL \cite{shu2018small}, which is carried out by training models to adapt to novel tasks with minimal data rapidly. Among meta-learning frameworks, Model-Agnostic Meta-Learning (MAML) \cite{finn2017model} has been widely adopted for ZSL due to its flexibility in optimizing model initialization for fast adaptation.

MAML seeks to find model parameters that are easy to adapt to new tasks using a small number of gradient steps \cite{finn2017model}. During meta-training, MAML simulates few-shot learning scenarios by splitting classes into meta-train and meta-test sets. For each episode, the model first updates its parameters on the support set (analogous to seen classes) and then evaluates on the query set (simulating unseen classes). This bi-level optimization process ensures the learned initialization captures cross-class transferable features, which is critical for zero-shot generalization.

However, its effectiveness is limited by small sample sizes and heterogeneous data distributions when applied to high-variance time series data. Several improvements to MAML have been proposed. A seminal adaptation of MAML to ZSL is proposed by \cite{Gidaris_2018_CVPR}, where attribute embeddings are used to condition the model's feature space. By meta-training on seen classes with their attributes, the model learns to adjust its decision boundaries for unseen classes based solely on their semantic descriptors. \citet{baik2021meta} proposed MeTAL (Meta-Learning with Task-Adaptive Loss), which learns task-specific loss functions to dynamically adapt to the characteristics of each task. Similarly, \citet{li2022meta} modified MAML by introducing subject-specific gradient updates and incorporating source-task knowledge to improve accuracy in such scenarios. 

Although meta-learning methods like MAML have shown potential in fast adaptation to new tasks, they face critical limitations in financial time series forecasting. 

The effectiveness of these methods heavily depends on task partitioning strategies during meta-training, where the existing partitioning strategies often overlook the unique characteristics of financial data. Random task partitioning, for instance, may fail to consider temporal correlations and disrupt the consistency of financial time series \cite{finn2017model}. \citet{mo2023few} proposed a more advanced approach that organizes similar subsequences into meta-tasks using DTW-based similarity matching. However, this method has two limitations. First, it is sensitive to noise and outliers, which can distort similarity measurements. Second, it focuses on global sequence alignment while missing important local temporal patterns \cite{song2022robust,waduge2021consensus}.

%% file: 3_method.tex
\section{Methodology}
In this section, we provide a detailed description of our proposed framework, specifically tailored for MAML-based meta-learning in zero-shot financial time series forecasting. We first formulate the problem and highlight the key challenge. Then, we present the overview of the entire framework. Lastly, we focus on the key components of the framework, including the multi-view embedding and the task construction module.

\subsection{Problem Formulation}

We consider a multivariate time series dataset \( X = \{ x_t \}_{t=1}^{T+H} \), where \( x_t \in \mathbb{R}^d \) represents the feature vector at time \( t \) with \( d \) features. The dataset is divided into two segments for pretraining and evaluation. Specifically, the pretraining dataset, denoted as \( \mathcal{D}_{\text{pretrain}} = \{ x_t \}_{t=1}^T \), consists of observations up to time \( T \) and is used to learn general temporal representations. The test dataset, \( \mathcal{D}_{\text{test}} = \{ x_t \}_{t=T+1}^{T+H} \), contains the subsequent \( H \) time steps, which are used to evaluate the model's ability to generalize to unseen patterns.

Our objective is to develop a predictive model \( f \) that maps the observed time series context to future values. Formally, the model \( f \) is defined as:
\begin{equation}
    f \colon \{ x_t \}_{t=1}^{T} \to \hat{X}_{T+1:T+H},
\end{equation}

where \( \{ x_t \}_{t=1}^{T} \) represents the input time series up to time \( T \), and \( \hat{X}_{T+1:T+H} \) denotes the predicted values for the next \( H \) time steps. The prediction can be either the stock price or the return, depending on the target variable of interest.

The training objective involves minimizing a loss function:
\begin{equation}
     \mathcal{L}(\hat{X}_{T+1:T+H}, X_{T+1:T+H}),
\end{equation}
where \( X_{T+1:T+H} \) are the true values of the stock return, and the loss quantifies the discrepancy between the predicted and ground truth values.

\begin{figure*}[htbp]
\centering
\includegraphics[width=\linewidth, keepaspectratio=true, height=0.3\textheight]{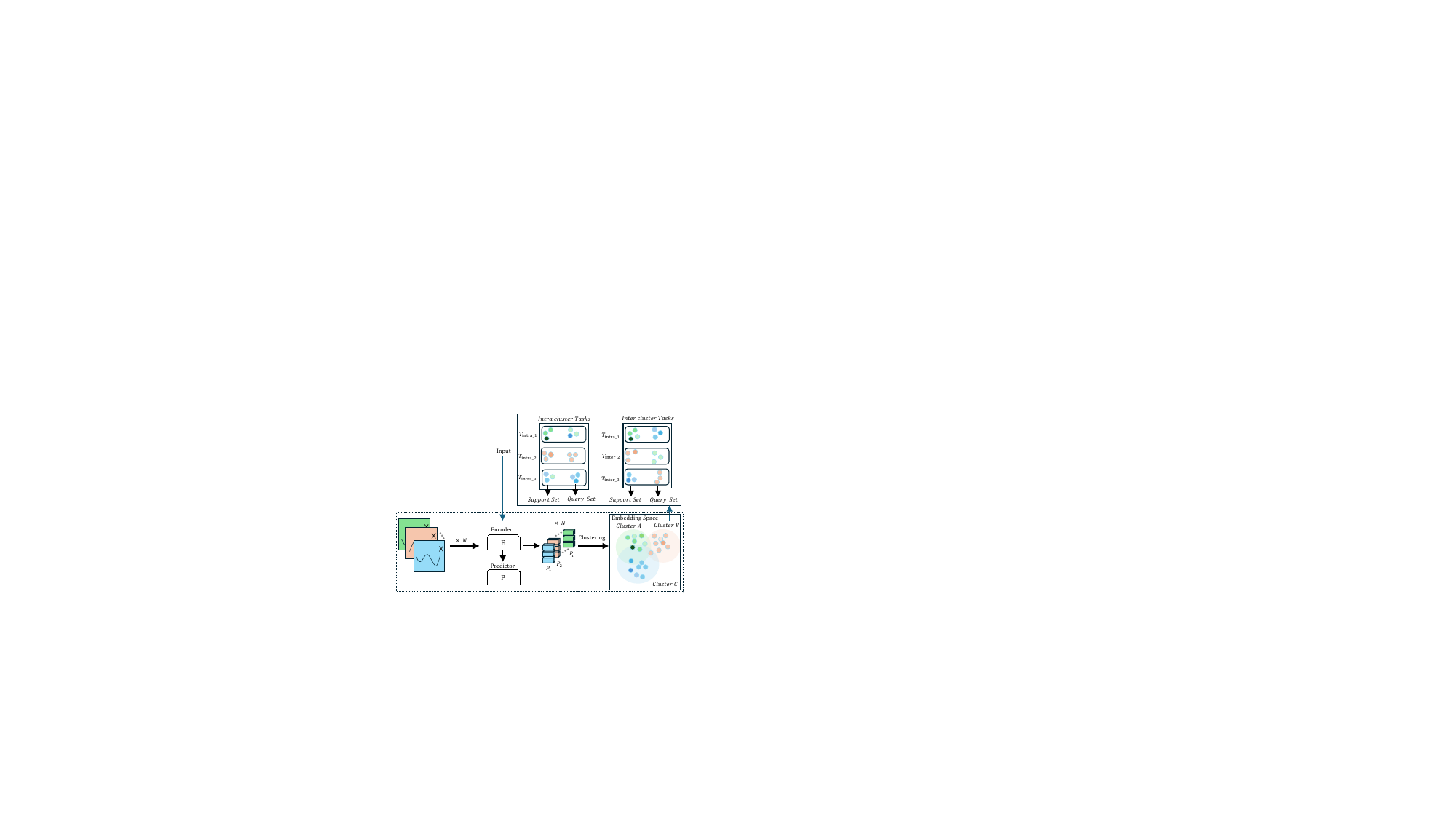}
\caption{Illustration of the proposed meta-learning framework. Input sequences are encoded and soft-clustered via GMM in the embedding space. Meta-tasks are constructed in three ways: (1) intra-cluster tasks sample both support and query sets from the same cluster; (2) inter-cluster tasks sample from different clusters; and (3) hard tasks pair support and query sets from the most dissimilar clusters. For clarity, only intra- and inter-cluster tasks are illustrated; hard tasks are not shown.}
\label{fig:meta_framework}
\end{figure*}

\subsection{Framework Overview}
As shown in Figure~\ref{fig:meta_framework}, we first randomly sample a batch of sequences (taken from the samples created via the sliding window strategy), denoted by size \(N\). We then feed these \(N\) sequences into the encoder \(E\), which produces batch embeddings of dimension \(N \times D\). 

Next, we apply Gaussian Mixture Models (GMM) for soft clustering on these embeddings, assigning each embedding probabilistically to \(K\) Gaussian components based on posterior probabilities. These probabilistic cluster assignments naturally capture the uncertainty and nuanced relationships inherent in temporal sequences.

Based on the soft clustering results, we construct two types of tasks, namely \emph{intra-cluster} tasks and \emph{inter-cluster} tasks. These tasks serve as meta-learning tasks, which are fed into the encoder and subsequently into a predictor to produce the final predictions. The full task construction procedure, including both intra-cluster, inter-cluster, and hard inter-cluster tasks, is detailed in Algorithm~\ref{alg:task_construction}.

\begingroup
\setlength{\baselineskip}{0.9\baselineskip}
{\small

\begin{algorithm}[!ht]
\caption{Embedding-based Task Construction}
\label{alg:task_construction}
\begin{algorithmic}[1]
\Input Encoder $E$, batch $\mathcal{X}$, number of components $K$
\Output Meta-tasks $\mathcal{T}$

\State $\mathbf{H} \gets E(\mathcal{X})$ \Comment{Extract sequence embeddings}
\State $\{\gamma_{ik}\} \gets$ GMM($\mathbf{H}$, $K$) \Comment{Soft clustering}

\State $\mathcal{T} \gets \{\}$

\For{$k = 1$ to $K$}
    \State Sample from component $k$ by $\gamma_{ik}$
    \State Split to support $\mathcal{S}_k$ and query $\mathcal{Q}_k$
    \State $\mathcal{T} \gets \mathcal{T} \cup \{(\mathcal{S}_k, \mathcal{Q}_k, \text{intra})\}$
\EndFor

\For{selected component pairs $(i, j)$}
    \State Sample support $\mathcal{S}$ from $i$, query $\mathcal{Q}$ from $j$ (by $\gamma_{ik}, \gamma_{jk}$)
    \State $\mathcal{T} \gets \mathcal{T} \cup \{(\mathcal{S}, \mathcal{Q}, \text{inter})\}$
\EndFor

\For{top dissimilar pairs $(i^*, j^*)$}
    \State Sample support $\mathcal{S}$ from $i^*$, query $\mathcal{Q}$ from $j^*$
    \State $\mathcal{T} \gets \mathcal{T} \cup \{(\mathcal{S}, \mathcal{Q}, \text{hard})\}$
\EndFor

\State \Return $\mathcal{T}$
\end{algorithmic}
\end{algorithm}
}
\endgroup

\subsection{Embedding Extraction}
\textbf{Base Encoder.} Let \(\mathbf{X} \in \mathbb{R}^{N \times L}\) be the input batch of \(N\) sequences (after any necessary preprocessing such as sliding windows), where each sequence has length \(L\). We denote our base encoder as \(E_\theta(\cdot)\), with learnable parameters \(\theta\). It can be any type of architecture, such as GRU, LSTM, or Transformer. The encoder transforms each sequence into a \(D\)-dimensional embedding. In matrix form, we can express the output of the encoder as
\begin{equation}
    \mathbf{H} = E_{\theta}(\mathbf{X}) \in \mathbb{R}^{N \times D},
\end{equation}
where each row \(\mathbf{h}_i \in \mathbb{R}^D\) corresponds to the embedding of the \(i\)-th sequence in the batch. The output embeddings are used in two ways: 1) task construction through clustering, where embeddings are used to group similar sequences together; and 2) prediction, where embeddings are fed into a predictor head for forecasting future values.

\textbf{Prediction Head.} The prediction head is implemented as a linear layer that maps the \({D}\)-dimensional embeddings to the next \({h}\) time steps. This transformation can be expressed as:
\begin{equation}
    \hat{\mathbf{y}} = \mathbf{W}\mathbf{h} + \mathbf{b},
\end{equation}
where \(\mathbf{W} \in \mathbb{R}^{h \times D}\) is the weight matrix, \(\mathbf{b} \in \mathbb{R}^h\) is the bias term, and \(\hat{\mathbf{y}} \in \mathbb{R}^h\) represents the predicted values for the next \(h\) time steps. 

\subsection{Tasks Construction}
After obtaining embeddings \(\mathbf{H}\), we apply Gaussian Mixture Models (GMM) for soft clustering to probabilistically partition embeddings into \(K\) Gaussian components. Each Gaussian component \(k\) is characterized by its mean vector \(\boldsymbol{\mu}_k\) and covariance matrix \(\boldsymbol{\Sigma}_k\), calculated as:
\begin{equation}
\boldsymbol{\mu}_k = \frac{\sum_{i=1}^N \gamma_{ik}\mathbf{h}_i}{\sum_{i=1}^N \gamma_{ik}}, \quad 
\boldsymbol{\Sigma}_k = \frac{\sum_{i=1}^N \gamma_{ik}(\mathbf{h}_i - \boldsymbol{\mu}_k)(\mathbf{h}_i - \boldsymbol{\mu}_k)^\top}{\sum_{i=1}^N \gamma_{ik}},
\end{equation}
where \(\gamma_{ik}\) represents the posterior probability that the embedding \(\mathbf{h}_i\) belongs to component \(k\).

Time-series data are often characterized by complex multi-modal distributions and smooth transitions between regimes, resulting in overlapping structures in the embedding space. GMM-based soft clustering naturally captures this uncertainty by allowing each embedding to have fractional membership across clusters, in contrast to hard clustering methods like K-means, which assign each point to exactly one cluster under the assumption of spherical cluster shapes. The flexibility of GMM (which models full covariance matrices for each component) therefore yields more diverse and representative cluster assignments. This enhanced cluster diversity in turn produces richer meta-learning tasks that reflect nuanced temporal relationships of real-world data.

\textbf{Intra-Cluster Task.} For each Gaussian component \(k\), we form an \emph{intra-cluster} task by sampling sequences based on their posterior probabilities \(\gamma_{ik}\). Concretely, we split the sampled sequences into a support set \(\mathcal{S}_k\) and a query set \(\mathcal{Q}_k\) (with \(\mathcal{S}_k \cap \mathcal{Q}_k = \varnothing\)). These tasks focus on a single cluster in the embedding space, enabling the model to adapt to the local cluster-specific distribution quickly. Since sequences within the same cluster share common temporal patterns, intra-cluster tasks help the model learn cluster-specific features effectively.

\textbf{Inter-Cluster Task.} To enhance generalization, we also construct inter-cluster tasks that span different Gaussian components. In a regular inter-cluster task, we randomly select two distinct components and sample the support set from one and the query set from the other. This setting forces the model to extract features that are invariant across clusters, as it must generalize knowledge from one cluster’s context to make predictions in another. To further challenge the model’s robustness, we create \emph{hard} inter-cluster tasks by deliberately choosing the most dissimilar component pair. We measure the dissimilarity between Gaussian components \(i\) and \(j\) using the symmetric KL divergence:

\begin{equation}
    d(\boldsymbol{\mu}_i, \boldsymbol{\Sigma}_i, \boldsymbol{\mu}_j, \boldsymbol{\Sigma}_j) = \frac{1}{2}\left[D_{KL}\left(\mathcal{N}_i \| \mathcal{N}_j\right) + D_{KL}\left(\mathcal{N}_j \| \mathcal{N}_i\right)\right],
\end{equation}

where \( \mathcal{N}_i = \mathcal{N}(\boldsymbol{\mu}_i, \boldsymbol{\Sigma}_i) \) denotes the multivariate Gaussian distribution with mean \( \boldsymbol{\mu}_i \) and covariance \( \boldsymbol{\Sigma}_i \).

We identify the most dissimilar component pair \((\hat{i}, \hat{j})\) by:
\begin{equation}
    (\hat{i}, \hat{j}) = \underset{1 \leq i \neq j \leq K}{\arg\max} \; d(\boldsymbol{\mu}_i, \boldsymbol{\Sigma}_i, \boldsymbol{\mu}_j, \boldsymbol{\Sigma}_j),
\end{equation}

For these hard tasks, we sample a support set \(\mathcal{S}_{\hat{i}}\) from component \(\hat{i}\) and a query set \(\mathcal{Q}_{\hat{j}}\) from component \(\hat{j}\). As these components are maximally separated, the model must handle vastly different distributions, which leads to the learning of robust and invariant associations rather than overfitting to cluster-specific patterns.

By combining intra-cluster tasks (which emphasize rapid adaptation to local patterns) with regular and hard inter-cluster tasks (which emphasize generalization across clusters), we create a curriculum that spans from easy to challenging adaptation scenarios. The number of intra-cluster, inter-cluster, and hard inter-cluster tasks in each batch is controlled by predefined ratios. This GMM-based task construction thus enriches the diversity and representativeness of the meta-tasks, enabling the model to generalize more effectively across varied time-series distributions.

\subsection{Learning Procedure}

\textbf{Meta Learning Procedure.}
We adopt a standard MAML-based meta-learning approach over the tasks derived from the clustering process \cite{finn2017model}. Let \(E_{\theta}(\cdot)\) be our base encoder, which serves dual purposes: generating embeddings for clustering and processing sequences during meta-learning. During clustering, we use \(E_{\theta}\) in inference mode to obtain embeddings that reflect the current model's understanding of predictive patterns. These embeddings then guide task construction through the clustering process described above.

For each meta-training iteration, we sample a batch of tasks comprising both intra-cluster and inter-cluster variants. Following MAML, each task \(\mathcal{T}\) is split into a support set \(\mathcal{S}\) and a query set \(\mathcal{Q}\). The support set is used for task-specific adaptation: we compute the task-adapted parameters \(\theta'_{\mathcal{T}}\) by taking one or few gradient steps on the support set loss. The query set then evaluates the adapted model's performance, and these query losses drive the meta-update of \(\theta\).

This dual use of the encoder creates a beneficial feedback loop: better embeddings lead to more meaningful task structures, which in turn help learn better embeddings through meta-training. As the encoder parameters \(\theta\) are updated, the quality of both the embeddings and the resulting task clusters improves. This creates a mutually reinforcing cycle that enhances the model's ability to generalize. 

\textbf{Learning Objective.}
For each meta-training task, model parameters are first adapted on the support set, and the meta-objective is evaluated on the query set. 
We employ the Pseudo-Huber loss as our primary loss for meta-learning due to its robustness against outliers, which commonly appear in time-series forecasting tasks. 
Let \(\hat{y}_i\) denote the predicted value and \(y_i\) denote the true value for sample \(i\) in the query set \(\mathcal{Q}\). The Pseudo-Huber loss for task \(\mathcal{T}\) is defined as:
\begin{equation}
    \mathcal{L}_{\mathcal{T}} = \frac{1}{|\mathcal{Q}|}\sum_{i=1}^{|\mathcal{Q}|}\delta^2\left(\sqrt{1+\left(\frac{y_i-\hat{y}_i}{\delta}\right)^2}-1\right),
\end{equation}
where \(\delta\) is a hyperparameter controlling the transition between quadratic and linear behaviors of the loss function.

When training over a batch of tasks, we aggregate these query losses and backpropagate to update parameters \(\theta\):
\begin{equation}
    \min_{\theta} \sum_{\mathcal{T} \in \text{Batch}} \mathcal{L}_{\mathcal{T}},
\end{equation}
where the loss for each task \(\mathcal{T}\) is computed on its query set after meta-adaptation.

%% file: 4_experiment.tex
\section{Experiments}

\subsection{Experimental Setting}

\begin{figure}[htbp]
    \centering
    \includegraphics[width=1.05\linewidth]{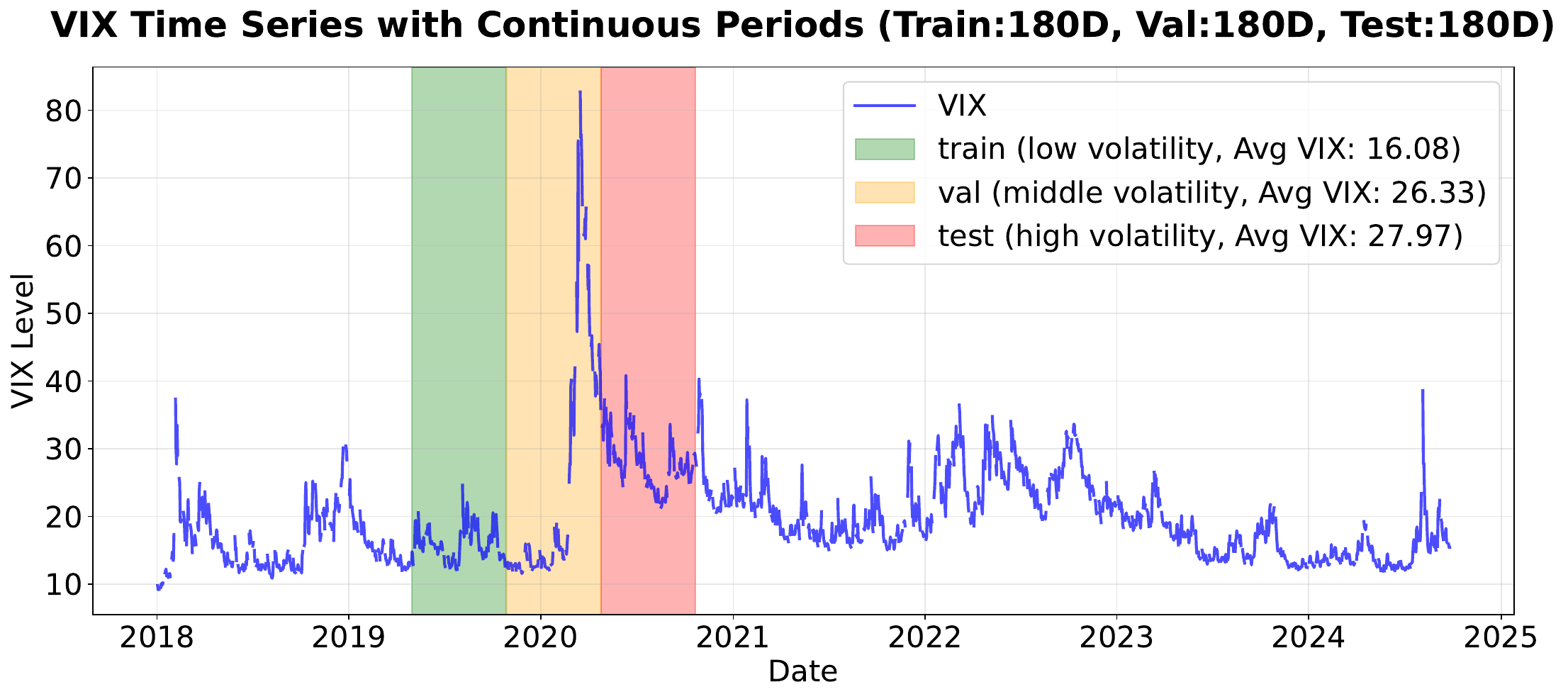}
    \caption{Data Split based on VIX}
    \label{fig:data_vix}
\end{figure}

\label{subsec:exp_setting}
\textbf{Datasets for Return Prediction.} To further validate our model's cross-market generalization capabilities between major markets and emerging economies, we utilized stock data from diverse sources: the S\&P 500 (USA), NSE (India), N225 (Japan), and HSCI (Hong Kong, China).

To evaluate the model’s performance under extreme market conditions, we focused on historical "black swan" periods. Utilizing the VIX index \footnote{Also known as the Chicago Board Options Exchange Volatility Index, which measures the market’s expectation of S\&P 500 volatility over the next 30 days and is often termed the "fear index". Higher VIX values typically indicate greater market panic and volatility} as our segmentation criterion, we calculated the average VIX value within every 180-day rolling window during 2018 to 2024. Based on these average VIX levels, we partitioned the dataset into three distinct, non-overlapping segments representing low, medium, and high volatility regimes: the training set covers April 1, 2019--- September 27, 2019; the validation set spans September 28, 2019---March 25, 2020; and the test set extends from March 26, 2020---September 21, 2020. Stocks were split into 60\%/20\%/20\% subsets for training, validation, and testing respectively, preventing any overlap in time windows or tickers. This strict separation in both time and stocks allows us to evaluate model performance under scenarios that mirror real-world challenges, such as newly listed stocks or significant distribution shifts due to market regime changes.

\begin{figure}[htbp]
    \centering
    \includegraphics[width=0.9\linewidth]{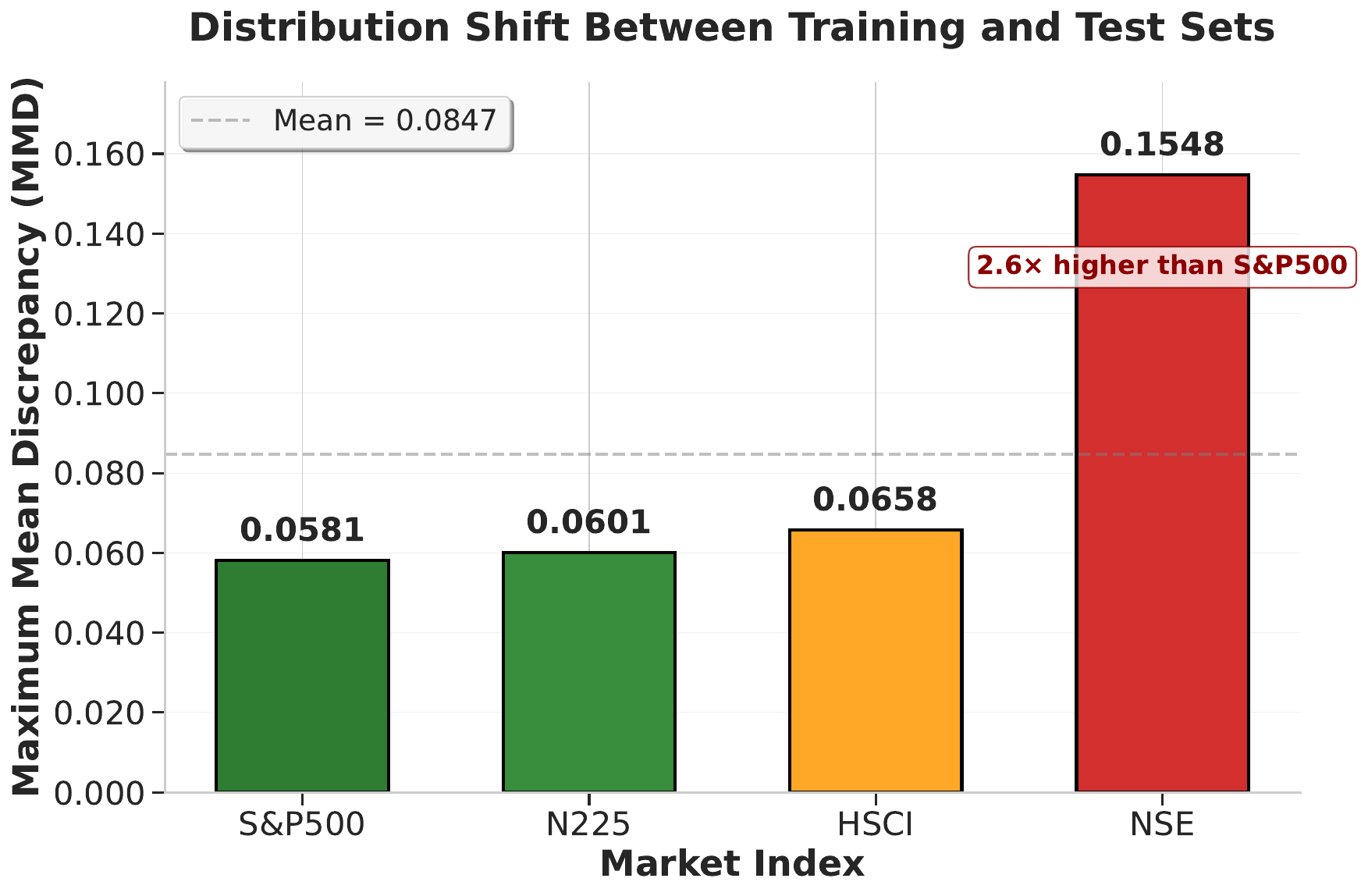}
    \caption{Maximum Mean Discrepancy (MMD) between training and test sets for each market.}
    \label{fig:mmd_shift}
\end{figure}

To quantify the effect of this split and assess the severity of train-test distribution shift under our experimental setup, we compute the Maximum Mean Discrepancy (MMD) between the training and test sets for each market after all partitions are finalized. As illustrated in Figure~\ref{fig:mmd_shift}, the US market (USA) exhibits the smallest distributional shift, while the Japanese market (JPN) and Hong Kong market (HKG) demonstrate moderate levels of shift. Notably, the distribution shift is most severe in the Indian market (IND). As an emerging economy, it is reasonable for the Indian market to experience more significant shifts when facing extreme market conditions, which aligns well with the objectives of our experiment.

Following \cite{zhao2023doubleadapt}, we construct input features using the \texttt{alpha360} toolkit from the Qlib platform \cite{lin2014qlib}. Due to the unavailability of VWAP data in our dataset, we adapt the feature generation process to create 300 technical factors based on the available price and volume metrics. We maintain the same 60\%/20\%/20\% stock split strategy to ensure no overlap in either temporal windows or ticker symbols. The target variable is defined as the next-day return relative to the current day (t+1/t). After that, we also apply cross-sectional normalization to both features and returns. 

\textbf{Evaluation Metrics.} For the return prediction task, we employ two categories of evaluation metrics. First, following \cite{lin2021learning, du2021adarnn, li2019individualized, xu2021hist, zhao2023doubleadapt}, we utilize four correlation-based metrics to evaluate the prediction accuracy: Information Coefficient (IC), Information Coefficient Information Ratio (ICIR), Rank Information Coefficient (Rank IC), and Rank Information Coefficient Information Ratio (Rank ICIR). Second, to evaluate the practical trading performance, we follow \cite{li2024master} to conduct a portfolio simulation on the test set. Specifically, we implement a daily trading strategy that selects the top 20 stocks with the highest predicted ranks and calculates two portfolio-based metrics: Excess Annualized Return (AR) and Information Ratio (IR). AR quantifies the annual excess return of our strategy, while IR measures the risk-adjusted performance of the investment strategy.

\textbf{Task‐Construction Baselines.} We consider the following methods as the comparison methods:
\begin{itemize}
    \item \textbf{No Meta-Learning (Pretrain)} employs a standard supervised learning approach where the base encoder is trained directly to predict price changes or returns without any meta-task structure.
    \item \textbf{Random-based Tasks}, following \cite{finn2017model}, constructs meta-tasks by randomly assigning samples within each batch of samples into k meta-tasks, where each task consists of a balanced split between support (50\%) and query (50\%) sets.
    \item \textbf{DTW-based Clustering}, inspired by \cite{mo2023few}, utilizes Dynamic Time Warping to compute similarity metrics between samples within each batch of samples. These similarity scores are then used to perform hierarchical clustering, resulting in distinct meta-tasks.
    \item \textbf{Euclidean distance-based Clustering}, following the approach in \cite{mo2023few}, applies Euclidean distance metrics to measure sample similarities within each batch of samples. Hierarchical clustering is then performed on these similarity measures to generate separate meta-tasks.
\end{itemize}

\textbf{Base Encoders.} All task‐construction methods employ three different encoders: GRU \cite{dey2017gate}, LSTM \cite{graves2012long}, and transformer \cite{vaswani2017attention}. This allows us to assess consistency across distinct neural architectures.

\textbf{Implementation Details.} We use a batch size of 512 and apply early stopping with a patience of 10 epochs, monitoring IC for return prediction. The sequence length is set to 3 for 1-day ahead prediction, since the input features have already aggregated information from the past 60 days, making longer sequences unnecessary. For the meta-task construction, we perform comprehensive hyperparameter optimization via random search over multiple parameters. The optimal configuration is selected based on validation set performance. 

\subsection{Return Prediction}
\label{subsec:return_exp}

\begin{table*}[!t]
\centering
\small
\begin{threeparttable}
\caption{Performance Comparison on NSE (India) and S\&P 500 (USA) Datasets. The best results in each dataset are in \textbf{bold} and the second best are \underline{underlined}.}
\label{tab:return_NSE_SP500}
\setlength{\tabcolsep}{3pt}
\renewcommand{\arraystretch}{0.9}
\begin{tabular}{ll|cccccc|cccccc}
\toprule
& & \multicolumn{6}{c|}{\textbf{NSE}} & \multicolumn{6}{c}{\textbf{S\&P 500}} \\
\cmidrule(lr){3-8} \cmidrule(lr){9-14}
\textbf{Model} & \textbf{Method} 
  & \textbf{IC} & \textbf{ICIR} & \textbf{RankIC} & \textbf{RankICIR} & \textbf{AR} & \textbf{IR} 
  & \textbf{IC} & \textbf{ICIR} & \textbf{RankIC} & \textbf{RankICIR} & \textbf{AR} & \textbf{IR} \\
\midrule
\multirow{5}{*}{GRU} 
 & Pretrain 
   & -0.0073 & -0.0765 & 0.0020 & 0.0205 & -0.0176 & 0.0380 
   & 0.0162 & 0.0943 & 0.0200 & 0.1018 & \underline{0.1734} & \underline{0.8820} \\
 & Random 
   & 0.0128 & 0.1418 & 0.0163 & 0.1748 & \underline{0.9986} & \underline{2.3225} 
   & 0.0033 & 0.0190 & -0.0027 & -0.0138 & 0.0893 & 0.6228 \\
 & DTW 
   & 0.0099 & 0.1097 & 0.0050 & 0.0510 & -0.0101 & 0.0556 
   & -0.0016 & -0.0079 & 0.0004 & 0.0018 & -0.0513 & -0.0484 \\
 & Euclidean
   & \underline{0.0174} & \underline{0.2063} & \underline{0.0194} & \underline{0.2476} & -0.0101 & 0.0556 
   & \underline{0.0186} & \underline{0.1025} & \underline{0.0237} & \textbf{0.1119} & 0.1298 & 0.7887 \\
 & \textbf{Proposed}
   & \textbf{0.0307} & \textbf{0.3608} & \textbf{0.0474} & \textbf{0.5280} & \textbf{1.1988} & \textbf{2.6292} 
   & \textbf{0.0259} & \textbf{0.1321} & \textbf{0.0209} & \underline{0.1022} & \textbf{0.5327} & \textbf{1.9622} \\
\midrule
\multirow{5}{*}{LSTM} 
 & Pretrain
   & -0.0415 & -0.4118 & 0.0131 & 0.1423 & \underline{1.4490} & \underline{3.2049} 
   & 0.0141 & \underline{0.0783} & 0.0118 & 0.0606 & \underline{0.1631} & \underline{0.8231} \\
 & Random
   & 0.0104 & 0.1213 & 0.0180 & 0.2047 & 0.2859 & 0.8360 
   & -0.0034 & -0.0178 & 0.0011 & 0.0053 & 0.1247 & 0.7422 \\
 & DTW
   & 0.0022 & 0.0252 & -0.0007 & -0.0075 & -0.4662 & -1.0667 
   & \underline{0.0143} & 0.0754 & \underline{0.0251} & \underline{0.1203} & 0.1187 & 0.6220 \\
 & Euclidean
   & \underline{0.0256} & \underline{0.2563} & \underline{0.0255} & \underline{0.2522} & 1.2393 & 2.7028 
   & 0.0118 & 0.0612 & 0.0145 & 0.0683 & -0.2082 & -1.1172 \\
 & \textbf{Proposed}
   & \textbf{0.0261} & \textbf{0.2878} & \textbf{0.0275} & \textbf{0.2995} & \textbf{1.6541} & \textbf{3.7441} 
   & \textbf{0.0251} & \textbf{0.1322} & \textbf{0.0295} & \textbf{0.1404} & \textbf{0.3175} & \textbf{1.6018} \\
\midrule
\multirow{5}{*}{\shortstack{Trans-\\ former}} 
 & Pretrain
   & 0.0029 & 0.0315 & 0.0069 & 0.0763 & 0.8297 & 1.8358 
   & -0.0264 & -0.1503 & -0.0278 & -0.1425 & -0.4534 & -3.1524 \\
 & Random 
   & 0.0142 & 0.1303 & 0.0167 & 0.1491 & 0.3290 & 0.7885 
   & 0.0049 & 0.0250 & 0.0123 & 0.0560 & -0.0027 & 0.0807 \\
 & DTW
   & \underline{0.0317} & \textbf{0.3494} & \underline{0.0301} & \textbf{0.3120} & \underline{2.1369} & \underline{4.0725} 
   & -0.0117 & -0.0565 & -0.0186 & -0.0814 & \textbf{0.2200} & \textbf{1.0515} \\
 & Euclidean
   & 0.0202 & 0.2336 & 0.0143 & 0.1726 & 0.6632 & 1.6137 
   & \underline{0.0085} & \underline{0.0449} & \underline{0.0175} & \underline{0.0845} & -0.1419 & -0.5560 \\
 & \textbf{Proposed}
   & \textbf{0.0347} & \underline{0.3399} & \textbf{0.0308} & \underline{0.2793} & \textbf{2.8039} & \textbf{5.0469} 
   & \textbf{0.0206} & \textbf{0.1082} & \textbf{0.0314} & \textbf{0.1490} & \underline{0.1048} & \underline{0.5933} \\
\bottomrule
\end{tabular}
\end{threeparttable}
\end{table*}

\begin{table*}[!t]
\centering
\small
\begin{threeparttable}
\caption{Performance Comparison on N225 (Japan) and HSCI (Hong Kong) Datasets. The best results in each dataset are in \textbf{bold} and the second best are \underline{underlined}.}
\label{tab:return_N225_HSCI}
\setlength{\tabcolsep}{3pt}
\renewcommand{\arraystretch}{0.9}
\begin{tabular}{ll|cccccc|cccccc}
\toprule
& & \multicolumn{6}{c|}{\textbf{N225}} & \multicolumn{6}{c}{\textbf{HSCI}} \\
\cmidrule(lr){3-8} \cmidrule(lr){9-14}
\textbf{Model} & \textbf{Method} 
  & \textbf{IC} & \textbf{ICIR} & \textbf{RankIC} & \textbf{RankICIR} & \textbf{AR} & \textbf{IR} 
  & \textbf{IC} & \textbf{ICIR} & \textbf{RankIC} & \textbf{RankICIR} & \textbf{AR} & \textbf{IR} \\
\midrule
\multirow{5}{*}{GRU} 
 & Pretrain 
   & -0.0095 & -0.0565 & -0.0126 & -0.0757 & 0.0041 & 0.0646 
   & 0.0075 & 0.0570 & 0.0013 & 0.0104 & 0.0644 & 0.4339 \\
 & Random 
   & 0.0256 & \underline{0.1443} & \underline{0.0302} & \textbf{0.1731} & -0.0336 & -0.4449 & -0.0141 & -0.0979 
 & -0.0158 & -0.1100 & -0.2125 & -1.4173 \\
 & DTW 
   & \textbf{0.0301} & \textbf{0.1531} & \textbf{0.0323} & \underline{0.1629} & \textbf{0.1854} & \textbf{2.0281} 
   & 0.0089 & 0.0597 & 0.0058 & 0.0391 & \textbf{0.7820} & \textbf{2.9144} \\
 & Euclidean
   & -0.0003 & -0.0018 & -0.0076 & -0.0426 & \underline{0.1732} & \underline{1.9675} 
   & \textbf{0.0210} & \textbf{0.1196} & \underline{0.0197} & \underline{0.1107} & 0.0435 & 0.2914 \\
 & \textbf{Proposed}
   & \underline{0.0265} & 0.1141 & 0.0202 & 0.0876 & 0.0706 & 0.6590 
   & \underline{0.0182} & \underline{0.1119} & \textbf{0.0203} & \textbf{0.1284} & \underline{0.2271} & \underline{1.0636} \\
\midrule
\multirow{5}{*}{LSTM} 
 & Pretrain
   & -0.0064 & -0.0358 & -0.0085 & -0.0460 & -0.0267 & -0.2062 
   & \underline{0.0057} & \underline{0.0461} & -0.0023 & -0.0174 & \textbf{0.1972} & \textbf{1.2308} \\
 & Random
   & -0.0126 & -0.0674 & -0.0135 & -0.0672 & -0.1001 & -0.9526 
   & 0.0030 & 0.0250 & \textbf{0.0047} & \textbf{0.0354} & 0.0020 & 0.0565 \\
 & DTW
   & \underline{0.0106} & \underline{0.0519} & \underline{0.0076} & \underline{0.0343} & -0.0076 & -0.1786 
   & 0.0050 & 0.0343 & -0.0012 & -0.0077 & -0.1165 & -0.6776 \\
 & Euclidean
   & -0.0202 & -0.1006 & -0.0237 & -0.1196 & \textbf{0.1501} & \textbf{1.6102} 
   & -0.0071 & -0.0487 & -0.0049 & -0.0332 & 0.1015 & 0.5626 \\
 & \textbf{Proposed}
   & \textbf{0.0172} & \textbf{0.0792} & \textbf{0.0113} & \textbf{0.0543} & \underline{0.1421} & \underline{1.3373} 
   & \textbf{0.0104} & \textbf{0.0571} & \underline{0.0009} & \underline{0.0045} & \underline{0.1924} & \underline{0.8313} \\
\midrule
\multirow{5}{*}{\shortstack{Trans-\\ former}} 
 & Pretrain
   & -0.0089 & -0.0495 & -0.0205 & -0.1158 & -0.0917 & -1.1471 
   & \underline{0.0006} & \underline{0.0043} & -0.0065 & -0.0439 & 0.0361 & 0.2124 \\
 & Random 
   & \textbf{0.0169} & \textbf{0.0912} & \underline{0.0099} & \underline{0.0516} & \underline{0.0904} & \textbf{1.1975} 
   & 0.0002 & 0.0012 & \underline{0.0045} & \underline{0.0276} & -0.1964 & -1.0727 \\
 & DTW
   & 0.0155 & 0.0816 & 0.0089 & 0.0442 & -0.0084 & -0.0569 
   & -0.0141 & -0.1022 & -0.0217 & -0.1525 & 0.0851 & 0.5001 \\
 & Euclidean
   & 0.0154 & \underline{0.0903} & 0.0013 & 0.0073 & -0.1013 & -1.0824 
   & -0.0077 & -0.0530 & -0.0161 & -0.1044 & \textbf{0.1504} & \textbf{0.8300} \\
 & \textbf{Proposed}
   & \underline{0.0166} & 0.0809 & \textbf{0.0170} &  \textbf{0.0838} & \textbf{0.1074} & \underline{1.1888} 
   & \textbf{0.0070} & \textbf{0.0435} & \textbf{0.0138} & \textbf{0.0808} & \underline{0.1197} & \underline{0.7080} \\
\bottomrule
\end{tabular}
\end{threeparttable}
\end{table*}


We present the performance comparison across different meta-learning task-construction methods on four distinct datasets: the NSE dataset, the S\&P 500 dataset, the N225 dataset, and the HSCI dataset. Overall, our proposed method consistently outperforms baseline approaches across various model architectures (GRU, LSTM, Transformer), highlighting its effectiveness in capturing meaningful temporal relationships through GMM-based task construction.

On the NSE dataset, our proposed method achieves the best results on all major metrics and models, with the sole exception of ICIR and RankICIR for the Transformer, where it ranks second, as shown in Table~\ref{tab:return_NSE_SP500}. Our framework consistently delivers the highest IC, AR, and IR for every model, and outperforms baselines by a clear margin across the vast majority of cases. These results highlight both the robustness and general applicability of our meta-learning approach.

For the S\&P 500 dataset, as shown in Table~\ref{tab:return_NSE_SP500}, our method achieves top performance across nearly all metrics and architectures, with only a few second-best scores. For instance, the GRU-based model improves IC from 0.0186 to 0.0259 and AR from 0.1734 to 0.5327. Meanwhile, the Transformer model increases ICIR from 0.0449 to 0.1082 and RankICIR from 0.0845 to 0.1490. These results further demonstrate the effectiveness and generalizability of our comprehensive meta-learning framework in challenging financial forecasting settings.

As shown in Table~\ref{tab:return_N225_HSCI}, our model demonstrates strong performance on the N225 index with both the LSTM and Transformer models, achieving the best or second-best score in most metrics. While its performance with the GRU model did not rank first (achieving only the second-best IC score), the overall results remain competitive, with minimal gaps to the top performers. This may stem from the exceptionally strong baseline performance of the GRU model itself in this particular market (outperforming both LSTM and Transformer), which makes our model's robust and generally effective feature extraction appear comparatively modest. In contrast, although the Random-Based Tasks method performs well here, its significant underperformance in other markets or models highlights its inherent randomness and lack of robustness.

For the HSCI index, our model demonstrates consistently strong performance across all three models, with the majority of metrics ranking first and only a small portion achieving second-best results.

Remarkably, under the extreme conditions observed across these four markets, our model consistently maintains positive returns, even achieving the best performance in most of the cases, notably outperforming other models which sometimes yielded negative scores. This robust performance underscores our model's strong stability and generalization capability, effectively adapting to unforeseen challenges in both volatile markets and emerging markets.

\begin{table}[H]
\centering
\caption{Ablation Study on GRU using the National Stock Exchange of India Dataset. The best results are in \textbf{bold} and the second best are \underline{underlined}.}
\label{tab:ablation_return_gru}
\setlength{\tabcolsep}{4pt}
\renewcommand{\arraystretch}{0.9}
\begin{tabular}{lcccc}
\toprule
\textbf{Method} & \textbf{IC} & \textbf{ICIR} & \textbf{RankIC} & \textbf{RankICIR} \\
\midrule
Vanilla          & 0.0094 & 0.1896 & 0.0235 & 0.2543 \\
+ Inter-cluster Tasks           & \underline{0.0216} & \underline{0.2484} & \underline{0.0237} & 0.2681 \\
+ Hard Tasks                     & 0.0125 & 0.1302 & 0.0146 & 0.1483 \\
\textbf{Proposed}      & \textbf{0.0299} & \textbf{0.3133} & \textbf{0.0328} & \textbf{0.3360} \\
\bottomrule
\end{tabular}
\end{table}

\subsection{Ablation Study}
\label{subsec:ablation}
To understand the individual contributions of each component in our method, we conduct an ablation study using GRU as the base encoder, evaluated on the NSE dataset. Table~\ref{tab:ablation_return_gru} shows the results, where each metric represents the average over five independent runs with different random seeds.

Starting from a basic embedding-based clustering method, adding inter-cluster tasks (CC) yields significant improvements, with IC rising from 0.0094 to 0.0216 and RankIC from 0.0235 to 0.0237, indicating that cross-cluster generalization notably benefits performance. However, incorporating only hard tasks (HT) without regular inter-cluster tasks produces comparatively limited gains and even underperforms the CC variant. This suggests that while hard tasks introduce valuable generalization challenges, they are most effective when combined with regular inter-cluster tasks.

The full proposed model, combining both inter-cluster and hard tasks, achieves the best results across all metrics, significantly surpassing each individual variant. The ICIR increases from 0.2484 (CC only) to 0.3133, and RankICIR from 0.2681 to 0.3360, clearly demonstrating the complementary nature of these components. These findings highlight the importance of integrating both moderate and highly challenging meta-task structures to enhance model robustness and generalization.

\section{Parameter Analysis}
\label{sec:param_analysis}

\begin{figure}[h]
    \centering
    \renewcommand{\arraystretch}{0.9}
    \includegraphics[width=0.85\linewidth]{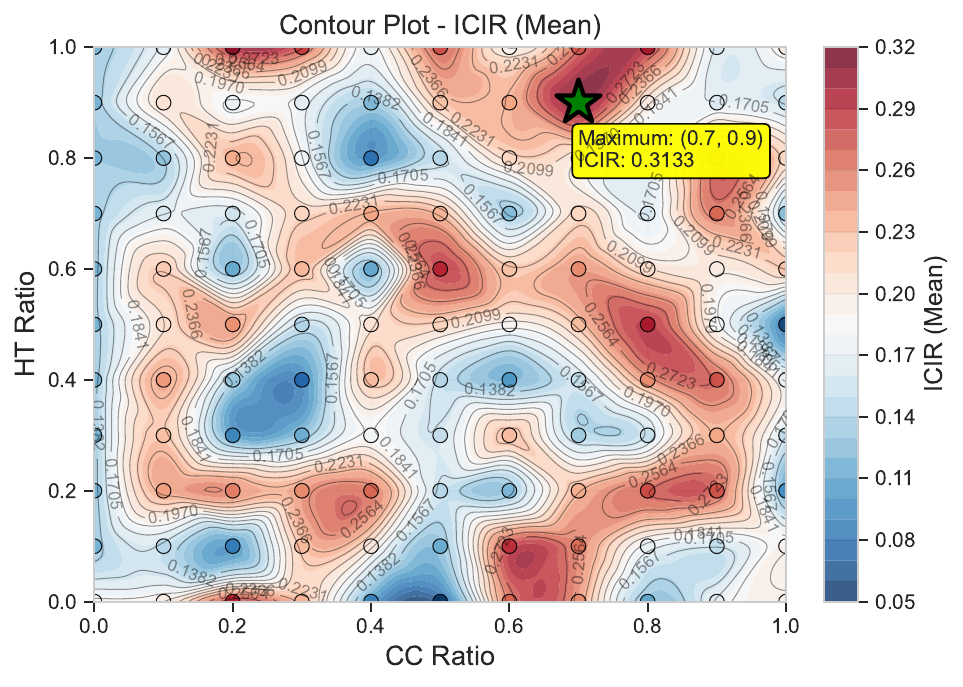}
    \vspace{0.2cm}
    \includegraphics[width=0.85\linewidth]{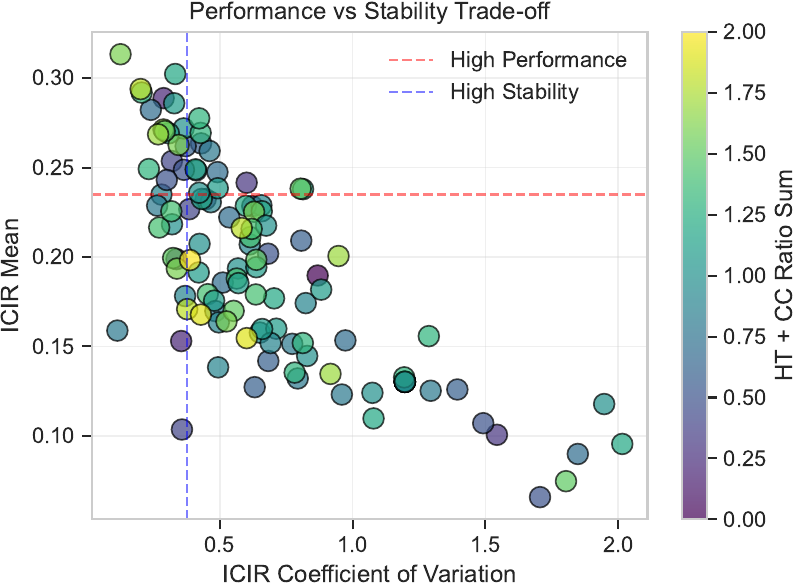}
    \caption{Top: Contour plot of ICIR (Mean) for the Indian market across HT/ICT parameter grid. Bottom: Trade-off between ICIR mean and coefficient of variation (CV) for different settings.}
    \label{fig:icir_param_sensitivity}
\end{figure}

To evaluate the robustness of our meta-task construction strategy, we conduct a comprehensive parameter sensitivity analysis via grid search over the hard task ratio (HT) and inter-cluster ratio (CC) on the Indian market dataset, which features pronounced distribution shifts. Given a fixed number of intra-cluster tasks per batch \(N_{\text{intra}}\), the number of inter-cluster tasks is \(N_{\text{intra}} \times \mathrm{CC}\), and the number of hard inter-cluster tasks is \(N_{\text{intra}} \times \mathrm{HT}\). Both HT and CC vary from 0.1 to 1.0, and each configuration is trained and evaluated five times with different random seeds. The results are visualized in Figure~\ref{fig:icir_param_sensitivity}.

Figure~\ref{fig:icir_param_sensitivity} (top) presents the contour plot of mean ICIR values across different CC and HT settings. The optimal parameter combination achieving the highest ICIR occurs at CC=$0.7$ and HT=$0.9$. However, the landscape is notably smooth and does not exhibit sharp, isolated peaks. Instead, multiple neighboring configurations, such as (CC=$0.8$, HT=$1.0$) and (CC=$0.8$, HT=$0.5$), yield comparably high ICIR values. This indicates that our proposed approach demonstrates robustness across a wide parameter region, rather than being sensitive to specific ratio settings.

\begin{table}[h]
    \centering
    \renewcommand{\arraystretch}{0.9}
    \caption{Most stable configurations (lowest CV for ICIR).}
    \begin{tabular}{cccc}
    \toprule
    CC & HT & CV (ICIR) & Mean ICIR \\  
    \midrule
    0.5 & 0.3 & 0.115 & 0.1589 \\     
    0.7 & 0.9 & 0.127 & 0.3133 \\
    0.8 & 1.0 & 0.202 & 0.2939 \\
    0.8 & 0.5 & 0.206 & 0.2919 \\
    0.7 & 0.6 & 0.233 & 0.2492 \\
    0.6 & 0.1 & 0.241 & 0.2823 \\
    0.4 & 0.4 & 0.264 & 0.2287 \\
    0.7 & 1.0 & 0.268 & 0.2686 \\
    0.6 & 0.8 & 0.272 & 0.2165 \\
    0.7 & 0.2 & 0.281 & 0.2349 \\
    \bottomrule
    \end{tabular}
    \label{tab:stable_configs}
\end{table}

Figure~\ref{fig:icir_param_sensitivity} (bottom) and Table~\ref{tab:stable_configs} show that strong and stable performance is generally associated with higher CC, while HT also plays a meaningful role across the top configurations. Many of the most stable settings have CC values of 0.5 or above, with a range of HT values including (CC=0.7, HT=0.9), (CC=0.8, HT=1.0), (CC=0.6, HT=0.1), and (CC=0.5, HT=0.3). This trend is also evident in the figure, where warmer-colored points are concentrated in the upper-left region. Although some extreme ratio configurations do not further improve stability and may even reduce it, the results suggest that maintaining a sufficiently high CC and including an appropriate proportion of hard tasks are both important for achieving robust and reliable performance. Properly tuning both CC and HT enables the meta-learner to benefit from diversity while maintaining stability.

%% file: 5_conclusion.tex
\section{Conclusion}
This paper presents a MAML-based meta-learning framework for zero-shot financial time series forecasting. Our approach uses a single encoder for both embedding extraction and meta-task construction, ensuring that improvements in representation learning and task construction reinforce each other throughout training. The GMM-based multi-level task construction, which includes intra-cluster, inter-cluster, and hard tasks, enables the model to adapt to fine-grained local patterns and generalize across market regimes. Extensive experiments demonstrate that our framework achieves superior performance on zero-shot and out-of-distribution forecasting tasks.  For future work, we aim to develop adaptive strategies for task ratio selection and explore more flexible sampling schemes beyond fixed sliding windows, further enhancing the effectiveness of our approach.